\documentclass{article}

\usepackage{arxiv}
\usepackage{hyperref} 
\usepackage[backend=bibtex,style=authoryear]{biblatex}

\usepackage[utf8]{inputenc} 
\usepackage[T1]{fontenc}    
\usepackage{url}            
\usepackage{booktabs}       
\usepackage{amsfonts}       
\usepackage{nicefrac}       
\usepackage{microtype}      
\usepackage{lipsum}
\usepackage{graphicx}
\usepackage{xcolor}
\usepackage{colortbl} 
\usepackage[ruled,vlined]{algorithm2e}

\graphicspath{ {./images/} }

\title{Randomness control and
reproducibility study of random forest algorithm in \texttt{R} and \texttt{Python}}

\author{
 Louisa Camadini \\
  \texttt{l.camadini@itm-stats.com} \\
   \And
 Yanis Bouzid \\
  \And
Maeva Merlet \\
  \texttt{m.merlet@itm-stats.com} \\
   \And
 Léopold Carron \\
  \texttt{leopold.carron@loreal.com} \\
}

\begin{document}

\maketitle

\begin{abstract}
When it comes to the safety of cosmetic products, compliance with regulatory standards is crucial to guarantee consumer protection against the risks of skin irritation. Toxicologists must therefore be fully conversant with all risks. This applies not only to their day-to-day work, but also to all the algorithms they integrate into their routines. Recognizing this, ensuring the reproducibility of algorithms becomes one of the most crucial aspects to address. 

However, how can we prove the robustness of an algorithm such as the random forest, that relies heavily on randomness? In this report, we will discuss the strategy of integrating random forest into ocular tolerance assessment for toxicologists. 

We will compare four packages: randomForest and Ranger (\texttt{R} packages), adapted in \texttt{Python} via the SKRanger package, and the widely used Scikit-Learn with the \textsf{RandomForestClassifier()} function. Our goal is to investigate the parameters and sources of randomness affecting the outcomes of Random Forest algorithms.  

By setting comparable parameters and using the same Pseudo-Random Number Generator (PRNG), we expect to reproduce results consistently across the various available implementations of the random forest algorithm. Nevertheless, this exploration will unveil hidden layers of randomness and guide our understanding of the critical parameters necessary to ensure reproducibility across all four implementations of the random forest algorithm. 
\end{abstract}

\section{Introduction}\label{intro}      

The role of cosmetics toxicological evaluation is to guarantee the safety of consumer products. This involves substantial investments in time and financial resources for product evaluation. To reduce these investments, machine learning is one of the keys to help toxicologists in decision-making. Over recent years, several models have been developed to compute risk assessment metrics.

In a safety context, machine learning is actively used to develop predictive models that can replace animal testing. Various methods have been developed over the past four decades to address increasingly complex needs such as general systemic toxicology or skin sensitization, as mentioned in \cite{burbank} and retrieved from \cite{eu09} and \cite{eu10}.
\vspace{\baselineskip}

To ensure that a new tool can be integrated into risk assessment strategy, the first and crucial step is the adoption by the community. Given the significant responsibility inherent in every cosmetic product safety report submitted to health authorities, this adoption process necessitates thorough testing of any new model that may be developed. If any doubts regarding the reproducibility of the algorithm's implementation arise, a toxicologist may request an evaluation of its impact on risk assessment. 

Upon initial observation, the use of random forests may seem unsettling. As of now, there is a lack of evidence regarding the reproducibility of this method across the most popular programming languages -- such as \texttt{R} and \texttt{Python} -- despite widespread discussions on performance comparisons as exemplified in \cite{marchese}. To bridge this gap, our study aims to compare four random forest packages in \texttt{R} and \texttt{Python}, thereby assessing the reproducibility of their outcomes. 
\vspace{\baselineskip}

The article will begin with a brief section on random forests, as understanding how they work is essential for revealing sources of randomness. This will be followed by a review of existing packages, then the methods for deciphering the differences between them and the associated results. At the end of the article, a clear procedure will enable us to achieve perfect reproducibility between the studied packages.

\section{Random forest algorithm}\label{RFalgo}      

The random forest algorithm combines the concepts of CART (Classification and Regression Trees) and \emph{bagging (bootstrap aggregation)}, whose concepts are fully detailed in \cite{biau}. 

Bagging uses the bootstrap method to train decision trees on different samples of the dataset, then aggregates their predictions to make a final decision. This approach improves model stability and robustness by reducing variance and over-fitting. However, it adds an extra layer of randomness to the construction of each tree, making model reproducibility challenging. 

In addition, the random forest introduces randomness by selecting only a subset of candidates features ($m$ candidates among the total number of features available $p$) at each node split, instead of considering all features. This random selection helps decorrelate the individual trees, making them more diverse and less prone to overfitting. This split is typically determined using a criterion, such as the Gini index to minimize:  
\begin{equation}
    Gini = 1 - \sum_{i=1}^{c} p_i^2
\end{equation}
$c$ is the number of classes in the classification problem and each $p_i$ represents the proportion of samples that belong to class $i$ (see \cite{daniya}).

Then the algorithm constructs a forest of decision trees, where each tree is trained on a bootstrap data sample and selects the best split among a random subset of features at each node. By averaging the predictions of multiple trees, random forests provide robust and accurate predictions. 
If necessary, Fig. \ref{summaryRF} in appendix provides a clear summary of this approach.

\vspace{\baselineskip}
Algorithm \ref{algo} describes how random forests work: 

\begin{algorithm}[H]
    \caption{Random forest} \label{algo}
    \KwIn{$x$ observation to predict}
    \KwIn{$D_{n \times p}$ data}
    \KwIn{$m \in \mathbb{N}$ number of candidate variables for node splitting}
    \KwIn{$B$ number of trees in the forest}
    \For{$k \gets 1$ \KwTo $B$}{
        \textbf{1.} Draw $\theta_k$ a bootstrap sample in $D_{n \times p}$\;
        \textbf{2.} Build a tree on $\theta_k$\;
        Each split is selected by minimizing the Gini impurity on a set of $m$ variables randomly drawn from the $p$\;
        This tree is denoted $h(.,\theta_k)$\;
    }
    \KwOut{$h(x) = \frac{1}{B}\sum_{k=1}^{B} h(x,\theta_k)$}
\end{algorithm}
\vspace{\baselineskip}

To summarize, up to now we have identified two main sources of randomness in the random forest algorithm, making model reproducibility challenging: 
\begin{enumerate} \label{SourcesOfRandomness}
    \item Bootstrap sampling: involves randomly sampling observations from the dataset. 
    \item Random selection of features: at each node split, the algorithm randomly selects a subset of $m$ candidates features to consider when looking for the best split. 
\end{enumerate}

\section{Existing Packages}\label{packages}      

We are going to study the characteristics of four packages implementing the random forest algorithm, two from \texttt{R} and two from \texttt{Python}, then compare them. 

The reproducibility of random forest models represents a significant challenge due to the variations between different implementations in \texttt{R} and \texttt{Python}. This discrepancy can be attributed to differences in the algorithms, randomization processes and optimization techniques used by the different software packages. Addressing this issue is essential to ensure the reliability and applicability of random forest models in the context of studies requiring compliance validations. 

\paragraph{\texttt{R} packages} 
\begin{itemize}
    \item \textbf{randomForest}: a standard \texttt{R} package that provides an implementation of the random forest algorithm, written in \texttt{C} and \texttt{Fortran}. \\
    \emph{Version:} \textsf{4.7.1.1}, \emph{function:} "\textsf{randomForest()}” 
    
    \item \textbf{Ranger}: an \texttt{R} package written in \texttt{C++}, providing a faster implementation. \\
    \emph{Version:} \textsf{0.12.1}, \emph{function:} "\textsf{ranger()}” 
\end{itemize}

\paragraph{\texttt{Python} packages} 
\begin{itemize}
    \item \textbf{SKRanger}: a \texttt{Python} compatible version of the Ranger \texttt{R} package, offering a Scikit-Learn interface and built on top of the Ranger library, which is written in \texttt{Python/Cython} and \texttt{C++}. Ranger and SKRanger share the same \texttt{C++} code, the difference being in the way it is called. \\
    \emph{Version:} \textsf{0.8.0}, \emph{function:} "\textsf{RangerForestClassifier()}”
    
    \item \textbf{Scikit-Learn}: a package primarily written in \texttt{Python} with some parts optimized using \texttt{Cython}. \\
    \emph{Version:} \textsf{1.1.2}, \emph{function:} "\textsf{RangerForestClassifier()}”
\end{itemize}

\paragraph{Parameter comparison} 
Table \ref{TableParams} summarizes the comparable function parameters of these packages and their purpose. Parameters in black are classified according to the documentation (\cite{sklearn.RandomForestClassifier}, \cite{skranger.RangerForestClassifier}, \cite{R.ranger} and \cite{R.randomForest}), but $nodesize$'s action is questioned in \ref{nodesize}, leading to the blue correction.

\begin{table}[t]
\begin{center}
\caption{Main parameters of the random forest functions in the four packages (\textcolor{blue}{purpose discussed in \ref{nodesize}})}
\label{TableParams}
\begin{tabular}{lcccc}
\hline
\rowcolor{gray!10} 
\textbf{Purpose} & \textbf{Scikit-Learn} & \textbf{SKRanger} & \textbf{Ranger} & \textbf{randomForest} \\
\hline
Number of trees in the forest & $n\_estimators$ & $n\_estimators$ & $num.trees$ & $ntree$ \\
\rowcolor{gray!20} 
Min node size to split at & $min\_samples\_split$ &$ min\_node\_size$ & $min.node.size$ & \textcolor{blue}{$nodesize$} \\
Min size of leaf node & $min\_samples\_leaf$ & & & $nodesize$ \\
\rowcolor{gray!20} 
Max depth of the tree & $max\_depth$ & $max\_depth$ & $max.depth$ & \\
Number of features to (randomly) \\ consider when splitting on nodes & $max\_features$ & $mtry$ & $mtry$ & $mtry$ \\
\rowcolor{gray!20} 
Bootstrap (sample with replacement) & $bootstrap$ & $replace$ & $replace$ & $replace$ \\
Size of sample to draw & $max\_samples$ & $sample\_fraction$ & $sample.fraction$ & $sampsize$ \\
\rowcolor{gray!20} 
Splitting rule & $criterion$ & $split\_rule$ & $split.rule$ & \\
\hline
\end{tabular}
\end{center}
\end{table}

\section{Methods}\label{Methods}      
To establish the common features and differences between Scikit-Learn, SKRanger, Ranger, and randomForest, we will implement a random forest model using each package and compare the outputs of the four models. 

SKRanger and Ranger share the same underlying \texttt{C++} code but interact with it differently. SKRanger utilizes \texttt{Cython} for \texttt{C++} function calls instead of \texttt{R}'s. As they both rely on the same Pseudo-Random Number Generator (PRNG), with a fixed random seed their outputs should be identical.  

\vspace{\baselineskip}
However, to ensure consistency in our experiments, we employed the $rpy2$ package, enabling the execution of \texttt{R} code directly within a  \texttt{Python} environment and facilitating comparisons. It is noteworthy that $rpy2$ utilizes the \texttt{R} kernel installed on the machine, meaning that \texttt{R}-based code executed through $rpy2$ will utilize \texttt{R}'s random number generator rather than \texttt{Python}'s.

\subsection*{Data} \label{Data}
Based on over a century of knowledge in cosmetics, L’Oréal has a huge history in assessing the risk of his product for local tolerance. To avoid any hazard to the consumer without using animal tests (which are not allowed since 2013), we have a wide variety of in vitro tests, giving each product an associated irritation score.  

Among this palette of in vitro tests, there is a method for assessing ocular irritation, which returns the irritation class of a formula. In this test, each formula is classified as \emph{non-irritant} (encoded 0), \emph{moderately irritant} (1) or \emph{severely irritant} (2). 

\vspace{\baselineskip}
We consider a dataset containing $N=4598$ rows corresponding to different formulas (tested product). A formula is the sum of ingredient concentrations, with each of the $p=87$ columns corresponding to a single ingredient contributing to the composition. The values indicate the percentage concentration in the formula, meaning that the sum of each row equals $100$.   

\vspace{\baselineskip}
The random forest procedure is used to predict the irritation class from the dataset. First, this dataset is split into training and testing sets, following the common practice of allocating $80\%$ of the dataset to the training set and reserving the remaining $20\%$ for testing, using the \textsf{train\_test\_split()} \texttt{Python} function. Results will be displayed on this test dataset, composed of $920$ rows corresponding to formulas and $87$ columns which are the ingredients. Train and test datasets are now fixed for the rest of the testing process.

\subsection{First implementation} 
For this first random forest implementation, we leave the default values of parameters to all four packages and run the random forest on $5000$ trees. To do so, only the parameters in Table \ref{Params} are fixed.
\begin{table}[b]
\begin{center}
\caption{Parameters set to be identical between packages (others take their default values)} \label{Params}
\begin{tabular}{ccccc}
\hline
\rowcolor{gray!10}
\textbf{Scikit-Learn} & \textbf{SKRanger} & \textbf{Ranger} & \textbf{randomForest} & \textbf{Value} \\
\hline
$n\_estimators$ & $n\_estimators$ & $num.trees$ & $ntree$ & \textbf{$5000$} \\
\hline
\end{tabular}
\end{center}
\end{table}

All other parameters are not specified meaning that default values apply, which may vary between packages. This results (in \ref{results1st}) will serve as a basis for comparing metrics and highlighting the need to control the randomness of algorithms, in addition to parameters.

\subsection{Eliminating sources of randomness} 
As mentioned in \ref{SourcesOfRandomness}, randomness is introduced at two stages in the model: if using bootstrap sampling, and at each node split when we randomly draw $m$ candidate variables from the $p$ variables in our data. For the four packages, these phenomena can be controlled by setting model parameters as in Table \ref{FixedParam}.

\begin{table}[b]
\begin{center}
\caption{Parameters set to manage randomness (others take their default values)} \label{FixedParam}
\begin{tabular}{ccccc}
\hline
\rowcolor{gray!10}
\textbf{Scikit-Learn} & \textbf{SKRanger} & \textbf{Ranger} & \textbf{randomForest} & \textbf{Value} \\
\hline
$n\_estimators$ & $n\_estimators$ & $num.trees$ & $ntree$ & \textbf{$1$} \\
\rowcolor{gray!20} 
$max\_features$ & $mtry$ & $mtry$ & $mtry$ & \textbf{$87$ (all)} \\ 
$bootstrap$ & $replace$ & $replace$ & $replace$ & \textbf{False} \\
\rowcolor{gray!20}
$max\_samples$ & $sample\_fraction$ & $sample.fraction$ & $sampsize$ & \textbf{$1$ (none)} \\
\hline
\end{tabular}
\end{center}
\end{table}

In this test, we will not perform bagging. Instead, models are trained on the entire train set defined beforehand, thus limiting the randomness of data allocation. All $p=87$ variables are considered at each node division (meaning all available variables).  

\vspace{\baselineskip}
\underline{Note}: Only one tree is needed, because considering all variables when dividing each node, and not doing bootstrap sampling would mean making the same decision tree every time.

\vspace{1.5cm}
\section{Results}\label{Results}      
All four models were pre-trained on the dedicated data, as explained above. This section presents the results of models' predictions on the previously defined test dataset. 

\subsection{First implementation} \label{results1st}
Table \ref{TableResults1st} shows the number of divergent classifications between each package on the test set, containing 920 observations (formulas).

\begin{table}[b]
\begin{center}
\caption{Number of divergent classifications between packages (\ref{results1st})}
\label{TableResults1st}
\begin{tabular}{r|cccc}
\hline
\rowcolor{gray!10} 
& \textbf{SKRanger} & \textbf{Ranger} & \textbf{randomForest} \\
\hline
\textbf{Scikit-Learn}   & $20$ & $5$ & $6$ \\
\rowcolor{gray!20} 
\textbf{SKRanger}   & & $19$ & $20$ \\
\textbf{Ranger}   & & & $1$ \\
\hline
\end{tabular}
\end{center}
\end{table}

These results and the similarities in the source code allow us to conjecture that the algorithm implementations of the randomForest, Scikit Learn and Ranger packages are broadly similar. However, discrepancies are observed in the outputs obtained with SKRanger. The reasons are not obvious, and the aim is to clarify them in order to achieve perfect reproducibility between packages. 

\subsection{Eliminating sources of randomness}  \label{results2nd}
In this test, no bootstrap is performed, and all variables are considered when searching for the best division. Table \ref{TableResults2nd} presents the number of divergent classifications.

\begin{table}[b]
\begin{center}
\caption{Number of divergent classifications between packages (\ref{results2nd})}
\label{TableResults2nd}
\begin{tabular}{r|cccc}
\hline
\rowcolor{gray!10} 
& \textbf{SKRanger} & \textbf{Ranger} & \textbf{randomForest} \\
\hline
\textbf{Scikit-Learn}   & $7$ & $8$ & $7$ \\
\rowcolor{gray!20} 
\textbf{SKRanger}   & & $3$ & $0$ \\
\textbf{Ranger}   & & & $3$ \\
\hline
\end{tabular}
\end{center}
\end{table}

Variations in classification remain between the four models. To better understand them, next section (\ref{treeanalysis}) is intended to analyse the trees returned by the models. In theory, only one tree is required in this context. 

\subsection{Analysis of generated trees} \label{treeanalysis}
In the following, we will impose a strict limit of 5 on the depth of trees, to ensure clear graphical representation. However, all observations concerning these trees have also been validated using trees without depth restrictions. Except for the depth, trees are generated as in \ref{results2nd} where sources of randomness are eliminated. Four fundamental results emerge from the tree-by-tree analysis and are presented in the following subsections.

\vspace{0.3cm}
\subsubsection{$1^{st}$ finding: Randomness in splitting variables drawing} \vspace{\baselineskip}
For the packages randomForest, Scikit-Learn and Ranger, generated trees are almost identical, but there are a few nodes for which the splitting variables are not. Indeed, by running the codes of these three packages several times, some splitting variables change. This is not the case for SKRanger, which uses a random seed set at 42 by default.  

\vspace{\baselineskip}
However, utilizing the Scikit-Learn package, which records tree details, we observe that while the splitting variables of specific nodes vary across executions, the Gini criterion of the node remains the same.  


This phenomenon is simply due to the random drawing of variables: although all variables are candidates in a node cut, they are drawn in random order. Thus, even if an equivalent cut is possible, it could be ignored, as a cut variable is only chosen if it strictly reduces the impurity criterion. 

\vspace{\baselineskip}
So, tree analysis reveals another source of randomness: for equal impurity criteria, the choice of variable splitting is random. Finally, by using the value of the Gini impurity criterion to compare the trees, the ones generated by the Scikit-Learn, randomForest and Ranger packages are identical.  

\vspace{0.3cm}
\subsubsection{$2^{nd}$ finding: $min\_node\_size$, a key parameter to set to ensure reproducibility with SKRanger} \vspace{\baselineskip} \label{minnodesize}

The tree generated by SKRanger has terminal node higher than the three other packages. Both Ranger and SKRanger use Ranger's \texttt{C++} code to build the trees, and with identical PRNG, the outputs should be perfectly similar. 

\vspace{\baselineskip}
With SKRanger, if a node contains less than 10 observations, it is not split. Research revealed that this was due to an unexpected overwrite of the $min\_node\_size$ default value (not fixed in \ref{FixedParam}), which is the minimum number of observations required in a node before it can be split.  Even though both \texttt{R} and \texttt{Python} packages rely on the same underlying \texttt{C++} library, apparently, they can independently adjust default parameter values. This flexibility leads to variations in the default $min\_node\_size$  parameter between implementations (set to 10 by SKRanger, 1 by others), resulting in discrepancies in output trees. 

To ensure efficient reproducibility between \texttt{R} and \texttt{Python}, it is advisable to set $min\_node\_size$ to $1$. Doing this, the trees generated by the four studied packages are now identical.

\vspace{0.3cm}
\subsubsection{$3^{rd}$ finding: Parameter $nodesize$ (randomForest) corresponds to the minimum node size to split at} \label{nodesize} \vspace{\baselineskip}

To ensure the robustness of the results, we systematically varied the model parameters, leading to another crucial discovery. 

According to the Scikit-Learn and randomForest packages documentation, the $min\_samples\_leaf$ and $nodesize$ parameters are defined as the \emph{“minimum size of leaf node”}: a split point at any depth will only be considered if it leaves at least the number of samples in each of the left and right branches. 

\begin{figure}[h] 
\begin{center}
\includegraphics[width=0.82\textwidth]{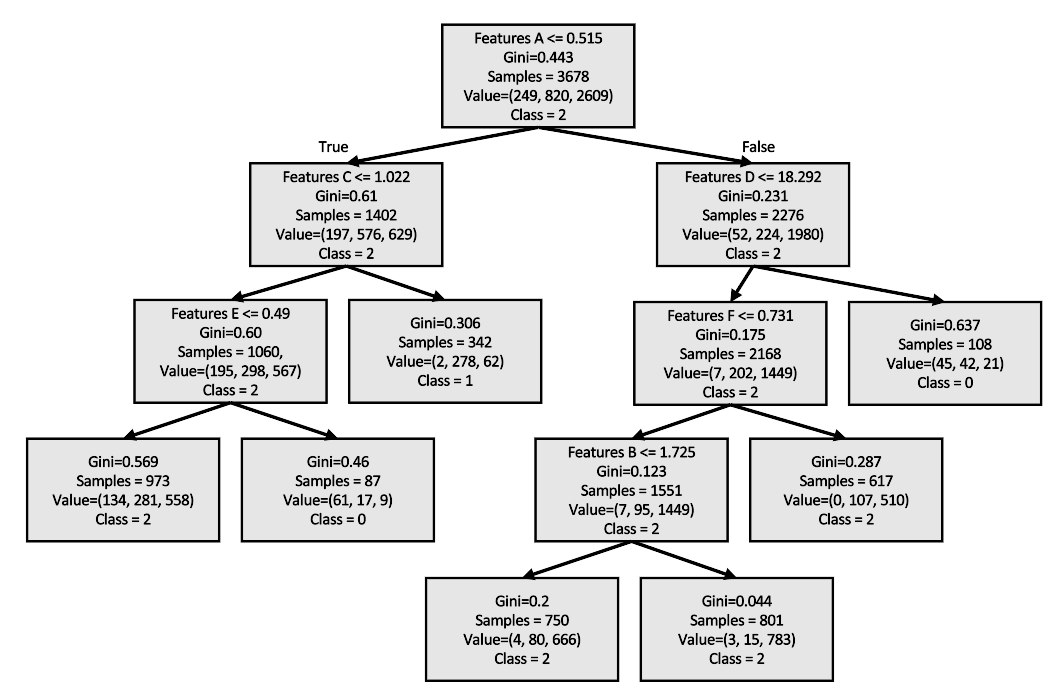}
\caption{Package randomForest, $nodesize=1000$}
\small Leaf nodes do not respect the minimum sample size set at 1000 by $nodesize$. \label{randomForest:nodesize=1000} 
\end{center}
\end{figure}

\begin{figure}[h] 
\begin{center}
\includegraphics[width=0.48\textwidth]{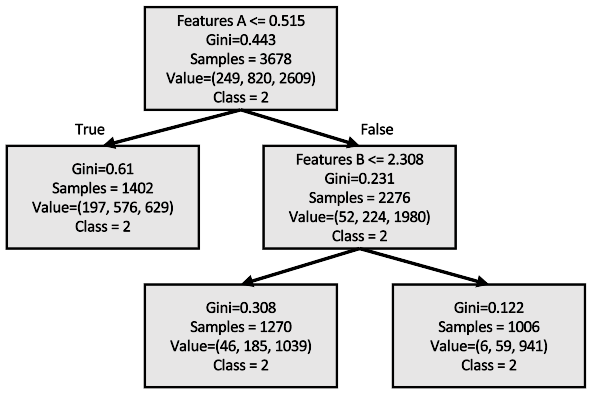} 
\caption{Package Scikit-Learn, $min\_samples\_leaf=1000$}
\small An example of how the $nodesize$ parameter should work, in theory equivalent to $min\_samples\_leaf$: leaf nodes both contain more than $1000$ samples.  \\
Nevertheless, when setting $min\_samples\_split=1000$, Scikit-Learn returns the same tree than in Fig. \ref{randomForest:nodesize=1000}.
\label{Scikit-Learn:min_samples_leaf=1000} 
\end{center}
\end{figure}

We have then compared the trees obtained by varying the parameters $nodesize$ of randomForest (Fig. \ref{randomForest:nodesize=1000}), as well as $min\_samples\_leaf$ and $min\_samples\_split$ of Scikit-Learn (Fig. \ref{Scikit-Learn:min_samples_leaf=1000}), with the reproducibility procedures explained so far (\ref{results2nd} and \ref{minnodesize}). 

\vspace{2cm}

Clearly, $nodesize$ acts as a $min\_samples\_split$, contradicting what is found in the documentation. Considering this, it is now possible to obtain similar trees for all four studied packages.

\vspace{2cm}
\subsubsection{$4^{rd}$ finding: Differences in aggregation methods (Ranger, SKRanger)} \vspace{\baselineskip}

Preceding steps make it possible to obtain identical trees by generating just one. This time, the aim is to extend the tree's reproducibility to the whole forest and compare the irritation score classifications on the test dataset containing $920$ rows. 

Nevertheless, when implementing random forests with both SKRanger and Ranger to predict the irritation class of the formulas – taking care to set the parameters just as before – this leads to $897/920$ equal classifications. 

\vspace{\baselineskip}
It turned out that SKRanger employs a different approach than Ranger: SKRanger does not rely on majority class voting mechanism to derive final predictions. Instead, it relies on predicting probabilities for each class through the \textsf{predict()} function. These probabilities represent the averaged values obtained by aggregating probabilities assigned to each class across all individual trees. In contrast, the majority class voting involves each tree \emph{voting} for a class, with the class receiving the most votes being chosen as the final prediction. Classifications may vary slightly between these two approaches, as discussed in greater detail in \cite{breiman}. 

\vspace{\baselineskip}
To match the predictions of both forests, it is possible to manually select the class associated with the highest probability for each Ranger prediction. When doing this, reproducibility is finally achieved by obtaining $920/920$ consistent classifications. 

It is worth noting that SKRanger (\textsf{v0.8.0}) uses Ranger (\textsf{v0.12.1}) code, suggesting possible implementation differences in newer Ranger versions (\textsf{v0.15.1}). 

\section{Discussion}\label{Discussions}      

The thorough analysis conducted confirmed the ability to generate identical decision trees across various implementations. However, locking in hyperparameters for reproducibility might not always be necessary in practical applications. Reproducibility is primarily a regulatory requirement, ensuring methodological integrity from input to output. 

\vspace{\baselineskip}
Although our approach is based on several tests on a single tree, all observations can be extended to a forest. While this starting point may not represent the most optimized implementation, the aim was to highlight the divergences between the four packages in a comprehensive way.

Similarly, some demonstrations have been made with a limited tree depth, in order to better understand the side effects that can modify the result between the two languages. As the difference in reproducibility can be observed at different depths, this choice does not affect our overall observation.

Therefore, we deliberately chose this comprehensive testing methodology to ensure the robustness and reliability of our analyses.

\vspace{\baselineskip}
In this article, we have delved into the \texttt{R} and \texttt{Python} implementations of random forests, and should emphasize two points.  
First, only \texttt{R} and \texttt{Python} packages are studied, whereas most articles comparing random forest performance usually include a third programming language. This choice stems from the fact that these are the two most widely used languages in data science.    

Second, we demonstrate how an internal effect of randomness can affect the overall robustness of an algorithm. Dealing with the reproducibility and robustness of randomness is a common problem in computer science, widely discussed in \cite{randomgen}. 

Despite the subtle differences we have found, they will have no impact on the use that most users have of random forests. This analysis focuses on reproducibility between several implementations and not on the internal reproducibility of a code, which can be dealt with by conventional guidelines \cite{peng}. 

From a broad perspective, it is clear that this small effect of variability will not be detected by a classical optimization of the model hyperparameter, and that most of the community has no need to evaluate the safety of their algorithm. With this analysis, our confidence in random forests is guaranteed for use in safety regulation, which was the objective of our research.  

\section{Conclusions}\label{Conclusions}      

In conclusion, achieving perfect reproducibility between Scikit-Learn, SKRanger, Ranger and randomForest packages involves several key steps.  

Firstly, common elements such as the random seed generator and environment parameters need to be standardized. 

Additionally, parameters with differing default values, such as the number of trees, should be set uniformly. Furthermore, meticulous adjustments are necessary to manage sources of randomness. This includes implementing models without bootstrapping and avoiding random variable selection during the search for the best split. Ensuring a consistent approach to sampling training and testing data is also essential. 

\vspace{\baselineskip}
Studying consistency at the individual level has revealed some crucial mechanisms. 
The choice of the splitting variable is random with equal impurity criterion, so based on the Gini criterion it is possible to obtain identical trees in all packages when taking special care to set $min\_node\_size$ to $1$ in SKRanger.  

Futhermore, the $nodesize$ randomForest parameter does not act as in the documentation, but as the minimum node size to split at. 

Finally, the reproducibility of  the entire forest is studied, by evaluating the number of matching classifications. This highlighted the differences between aggregation methods, providing further insights on why overall classifications may still slightly vary between packages. 

\section*{Acknowledgments}
The authors would like to thank L'Oréal for enabling us to carry out this research. Special thanks to Marie Thomas, Caroline Rouas, Elise Warcoin and Sophie Loisel Joubert for paving the way for this project. Another important person to mention is Luc Souverain for encouraging us to publish this article.

\section*{Author contribution}
Y.B produced all the analyses, Le.C lead the project, Lo.C, Le.C and M.M wrote the paper. 

\printbibliography

\end{document}